# Perspective-Corrected Spatial Referring Expression Generation for Human–Robot Interaction

Mingjiang Liu, Chengli Xiao, and Chunlin Chen, *Senior Member, IEEE*

***Abstract*—Intelligent robots designed to interact with humans in real scenarios need to be able to refer to entities actively by natural language. In spatial referring expression generation (REG), the ambiguity is unavoidable due to the diversity of reference frames, which will lead to an understanding gap between humans and robots. To narrow this gap, in this article, we propose a novel perspective-corrected spatial REG (PcSREG) approach for human–robot interaction (HRI) by considering the selection of reference frames. The task of REG is simplified into the process of generating diverse spatial relation units. First, we pick out all landmarks in these spatial relation units according to the entropy of preference and allow its updating through a stack model. Then, all possible referring expressions are generated according to different reference frame strategies. Finally, we evaluate every expression using a probabilistic referring expression resolution model and find the best expression that satisfies both of the appropriateness and effectiveness. We implement the proposed approach on a robot system and empirical experimental results show that our approach can generate more effective spatial referring expressions for practical applications.**

***Index Terms*—Human–robot interaction (HRI), perspective-taking, reference frame, spatial referring expression generation (REG).**

## I. Introduction

HUMAN–ROBOT interaction (HRI) has been viewed as an important domain of robotics and artificial intelligence. Generally speaking, HRI can be classified into four interaction paradigms, i.e., robot as a tool, robot as a cyborg extension, robot as an avatar, and robot as a sociable partner [1]. Except for some special purpose [2], there are a growing number of applications for robots to act as sociable partners, such as elderly care [3], children education [4], and entertainment [5]. HRI has attracted a large number of researchers to devote to make robots be qualified collaborators, assistants, or partners [6]–[8]. As a partner shares the same environment with humans, it is an essential skill to be able to refer to entities. Like a tour guide, he should refer to the destination for tourists; while as a rescuer, he needs to accurately describe the rescue site.

Manuscript received February 6, 2022; accepted March 14, 2022. This work was supported in part by the National Natural Science Foundation of China under Grant 71732003 and Grant 62073160, and in part by the National Key Research and Development Program of China under Grant 2018AAA0101100. This article was recommended by Associate Editor X. Wu. *(Corresponding author: Chunlin Chen.)*

This work involved human subjects or animals in its research. Approval of all ethical and experimental procedures and protocols was granted by the Ethics Committee of the Department of Psychology, Nanjing University under Application No. NJUPSY201904006.

Mingjiang Liu and Chunlin Chen are with the Department of Control and Systems Engineering, School of Management and Engineering, Nanjing University, Nanjing 210093, China (e-mail: clchen@nju.edu.cn).

Chengli Xiao is with the Department of Psychology, School of Social and Behavioral Sciences, Nanjing University, Nanjing 210023, China (e-mail: xiaocl@nju.edu.cn).

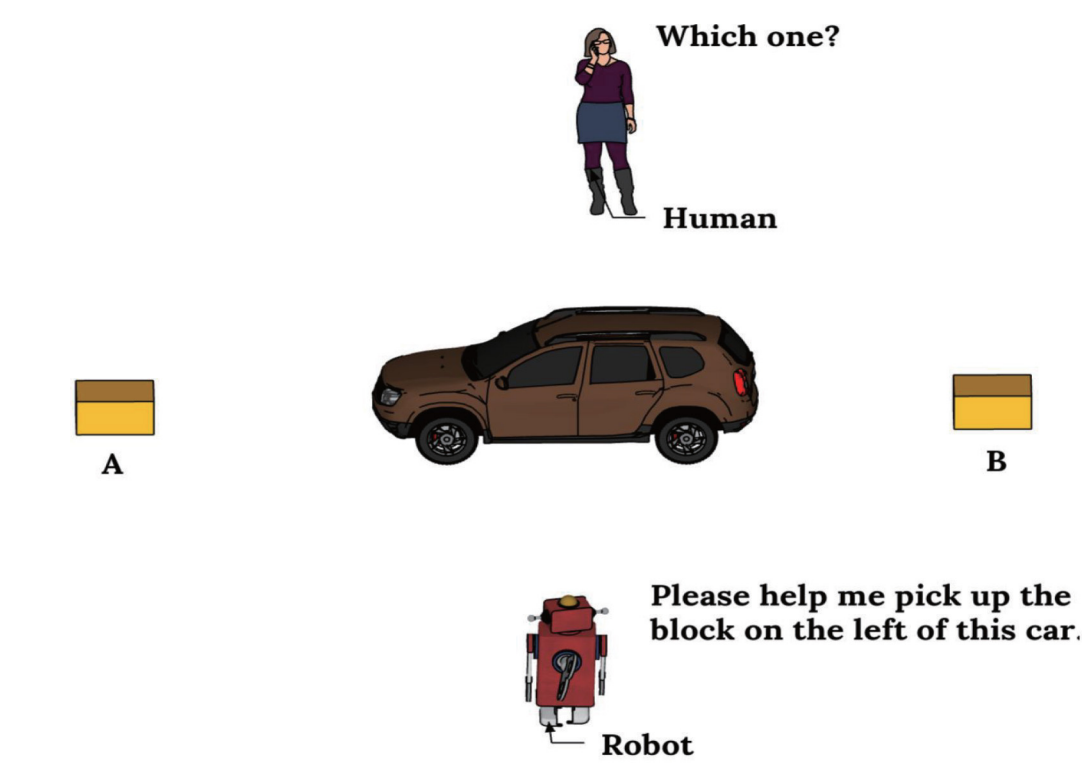

Fig. 1. Situation of HRI and we can clearly find the possible ambiguity of spatial referring expressions.

There are various ways to refer to entities, such as hand gestures and eye gaze; while natural language is most frequent and natural for human partners. The ability for robots to refer to entities by natural language is well known as referring expression generation (REG) [9]. In general, REG is concerned with how we produce a description of an entity that enables the hearer to identify that entity in a given context [10], which can help the robots to automatically perceive and describe the shared environment rather than just following the instructions passively [11], [12]. In this article, we aim at developing a robot system that can generate effective referring expressions for human partners in real-world scenarios.

Consider an interactive situation shown as in Fig. 1, where the referenced entities are material objects (e.g., blocks and cars) rather than abstract entities (e.g., places [13] and events). To refer to entities clearly, we usually use visual features (e.g., "the block") to describe the target entity. The expression using visual features may be insufficient, for example, there are likely to have two blocks in the domain. According to the human experience, one crucial solution is to utilize spatial references to represent the spatial relationship between two entities, e.g.,

"the block on the left of the car." Unfortunately, even with the use of spatial references, it is still possible to encounter additional ambiguity because of the uncertainty of "left" for that the spatial preposition "left" depends on the speaker's perspective or that of the others. This ambiguity originates from the selection of reference frames [14] and the spatial reference is a qualitative spatial location of objects concerning the selected reference frame. Different reference frames lead to different spatial perspectives and the description varies with the selected reference frame. Hence, when humans and robots select different reference frames, there will be an interactive gap between them due to ineffective referring expression.

To narrow this gap, more efforts are necessary in the design of human-centered robots. Although it is equivalent for robots to select any reference frame, we should attach importance to the human mental model. In other words, selecting a particular reference frame to resolve or generate referring expression is a natural cognitive activity of human and we should fully take into account the reference frame selection preference of human when designing robots.

In this article, we propose a novel perspective-corrected spatial REG (PcSREG) approach that takes the reference frames into account. Specifically, we simplify the REG task to the discrete process of generating basic relation units with the assumption that the real preference of reference frames can be observed. Instead of generating a complete referring expression for the target entity, we pick out all landmarks that can help to locate the target entity following a default reference frame and to generate corresponding visual features. Then, we generate spatial prepositions with a selected specific reference frame. Hence, our approach devotes to choose the most reasonable reference frame for each spatial relation unit. To formulate the human mental model, we propose a probabilistic referring expression resolution model to imitate the parsing process of humans. Besides, we propose a multiobjective joint optimization function and apply an exhaustive search strategy to find the best reference frame strategy (i.e., the best combination of reference frames). However, the assumption that we can fully understand the characteristics of human cognition is the biggest obstacle to implement the proposed approach and we provide an iterative mechanism to update the preference probability.

In this article, we implemented PcSREG on a robotic system and contribute a study to estimate the preference of reference frames in a tabletop task, where the preference reflects the cognitive characteristics of the resolved and generated human spatial expressions. The empirical experiments show that the PcSREG approach can generate more effective spatial referring expressions for HRI. The main contributions of this article are threefold.

1) We present a novel REG approach to generate perspective-corrected spatial referring expression, which proves to be a successful attempt to reduce the ambiguity that originates from reference frames.
2) We contribute a psychological study on the selection preference of reference frames in the generation and comprehension of human natural language.
3) We implement the proposed PcSREG approach on a robotic system and show the effectiveness of the whole system for HRI.

The remainder of this article is organized as follows. After discussing related work in Section II, we systematically introduce the proposed PcSREG approach in Section III, including the framework and specific details of algorithm design. In Section IV, we implement PcSREG on a robotic system. In Section V, the preference of reference frames is estimated and empirical experimental results demonstrate the success of the proposed approach. Conclusions are given in Section VI.

## II. Related Work

### A. Perspective Taking for HRI

Perspective taking is a cognitive activity of human to think or act by putting himself/herself in an alternative point of view. It is clear that the perspective-taking abilities of robots would be a valuable asset for HRI. In a narrow sense, perspective taking for HRI refers to that a robot should rotate its view to a person's perspective, where the "perspective" is just used to represent the mental model or physical view of the human or the robot. There have been some practices to integrate the notion of perspective taking into the design of robots. For example, a geometric reasoner is presented to generate symbolic relations from the perspectives of human and robot [15]. The task of robots is planned and the configurations are generated for the robot by thinking from the human's perspective [16], [17]. More attention has been paid to visual perspective taking, which is the ability to predict the visual experience of another individual in physical space [18]. Specifically, there are two levels in visual perspective taking, i.e., 1) predict whether another individual can see an object or not and 2) understand how an object is organized from another individual's point of view. In [19], level 1 visual perspective taking is performed on the robotic platform Coyote via the cognitive architecture Polyscheme, and reasoning is realized through comprehensive analysis from the perspectives of human and robot. Level 1 visual perspective taking is also used to help robots remove ambiguities [20]. Psychological evidence shows that the two levels of perspective taking depend on two qualitatively different computational processes, respectively [18]. Inspired by this study, a line of sight tracing algorithm was implemented for level 1 perspective taking and a mental rotation algorithm of point cloud was employed for level 2 perspective taking in unconstrained environments [21].

In a broad sense of perspective taking, the "perspective" is not only used to represent the views of human and robot, but also to represent other reference frames. In this sense, the "perspective" is called as "spatial perspective" that refers to the physical point of view reflected in a qualitative spatial location description of objects [22], and it has the same meaning as reference frames. In this article, we focus on this broad meaning of perspective taking and name it as "spatial perspective taking," which can be replaced by "reference frame selection." For spatial perspective taking, the existing work mainly focuses on human–human interaction [14], [22], [23] and a few works endow robots with the perspective-taking ability. Recently, a robot system was implemented to attempt to handle spatial perspectives reliably [24]. It can deal with simple scenes by associating a set of possessive keywords (e.g., "my" and "your") regarding each perspective, while it is grounded for reference expressions rather than generating.

### *B. Referring Expression Generation*

REG is a classic task that has been widely studied in natural language generation [25], whose most fundamental function is to produce a description for a particular entity. In early stage, REG has been formulated as a search problem that determines a set of features or relations to identify a target entity from distractors in the domain [26], and some typical algorithms can be applied, including the full brevity algorithm [27], the greedy heuristic algorithm [27], and the incremental algorithm [28]. These algorithms are simple and efficient for computational models of referring, but they reduce effectiveness to the uniqueness and rest on a number of simplifications of the actual REG task: the target is always just one object, not a set; the properties of objects are always crisp, never vague; objects are assumed to be equally salient and they never discuss linguistic realization. To link REG with a more generic mathematical formalism, the graph-based algorithm (GBA) for REG was proposed [29]. GBA formalizes a scene as a labeled directed graph, describes content selection as a subgraph construction problem, and proposes cost functions to guide the search process and to give preference to some solutions over others. The combination of graph theory and cost functions paves the way for a natural integration of rule-based generation techniques with stochastic approaches. To improve the practicality of REG, a hypergraph-based approach was proposed for referring to target sets [30]. The reference expression can reference spatial relations between objects effectively [31], [32]. To generate human-like expressions, GBA is extended by modeling user's individual variation in overspecification [33]. In addition, the mismatched perceptual capabilities between humans and robots in the real world was addressed through various ways [11], [30], [34]. Viewing referential communication as a collaborative process, dynamic description strategies are proposed in [34] and [35]. Based on a semantic knowledge base called as ontology, a domain-independent REG approach is proposed to generate unambiguous referring expressions and achieved state-of-the-art performance [12]. Unfortunately, most of the existing work does not consider the impact of reference frames in REG; while in situated dialogue, the partners can take various spatial perspectives to generate and understand the expression.

Inspired by recent successes of deep learning methods for image captioning, some models devoted to generating expressions in more complex environments [36]–[38]. Some methods were proposed to generate and comprehend object expressions simultaneously [39], [40]. Although these models are more applicable to complex situations, they are limited in the image space and generate reference expressions following a default perspective all the time, which ignores the diversity of reference frames in actual interaction. It is crucial to choose a reasonable reference frame for generating effective spatial reference expressions [14]. Following this guideline, an algorithm was designed to generate named spatial references [13]. This initial algorithm normally selects the reference frames that humans also select. But it is applied to the locations of geographic scale which does not work in small-scale elaborate interactive scenes (e.g., the tabletop task) for HRI.

TABLE I
NOTATIONS

| Notations | Descriptions |
|---|---|
| $\mathcal{O}$ | The set of the robot's world model of entities, $\mathcal{O} = \{o_1, o_2, \cdots, o_d\}$ |
| $\mathcal{O}_{cl}$ | The set of candidate landmarks |
| $o_t$ | The intended target entity, $o_t \in \mathcal{O}$ |
| $\mathcal{O}_{dt}$ | The set of distractor objects |
| $o_{dt}$ | A distractor object, $o_{dt} \in \mathcal{O}_{dt}$ |
| $o_{target}$ | The target to be referenced in a spatial relation unit |
| $o_l$ | The landmark in a spatial relation unit |
| $\mathcal{R}$ | The set of spatial prepositions, $\mathcal{R} = \{r_1, r_2, \cdots, r_m\}$ |
| $r_{sp}$ | The spatial preposition used in a spatial relation unit |
| $\mathcal{F}$ | The set of reference frames, $\mathcal{F} = \{f_j\}$, $j = 1, 2, \cdots, n$ |
| $\mathrm{RE}$ | The referring expression generated by robots |
| $k$ | The complexity of the generated $\mathrm{RE}$ |
| $\tau$ | The reference frame strategy |
| $p_{f_j}$ | The estimated preference of the reference frame $f_j$ |
| $H$ | The entropy of the preference distribution |
| $p_{man}(o\|\mathrm{RE})$ | The probability that the human will resolve $\mathrm{RE}$ to the object $o$ |
| $p_{rob}(\mathrm{RE}\|o_t)$ | The probability that the robot generates spatial description $\mathrm{RE}$ for target $o_t$ |
| $\omega_1(\mathrm{RE})$ | The appropriateness of $\mathrm{RE}$, $\omega_1(\mathrm{RE}) \in \{0, 1\}$ |
| $\omega_2(\mathrm{RE})$ | The effectiveness of $\mathrm{RE}$, $\omega_2(\mathrm{RE}) \in [0, 1]$ |
| $Relation(\langle o, o' \rangle)$ | The spatial relationship between object $o$ and $o'$ |
| $D_{vf}$ | The set of visual features used to describe the landmarks and target in a $\mathrm{RE}$ |
| $d_{vf}$ | The visual features used to describe an object |
| $D_{sp}$ | The set of spatial prepositions used to identify the target $o_t$ |
| $Stack$ | The stack model |

## III. PERSPECTIVE-CORRECTED SPATIAL REFERRING EXPRESSION GENERATION

In this section, we first give the problem formulation and then the proposed PcSREG approach is presented in detail with related techniques and specific algorithms.

### *A. Problem Formulation*

The ambiguity of referring expression caused by the diversity of reference frames often exists in real scenarios. As an example, imagine that a robot and a person are interacting in the situation shown as in Fig. 1. When the expression "the block on the left of this car" is uttered by the robot, it will make the human partner confused, because each of the two yellow blocks can be identified. This ambiguity origins from the diversity of reference frames. Different reference frames represent different spatial perspectives, block *A* will be identified when the person takes the perspective of the robot while block $B$ will be identified when the person takes the perspective of herself. It is clear that, the intended target object will be mistakenly identified when there is a mismatch in taking perspectives. We call this phenomenon as the ambiguity of spatial referring expressions. In this article, we focus on how to avoid or alleviate this ambiguity.

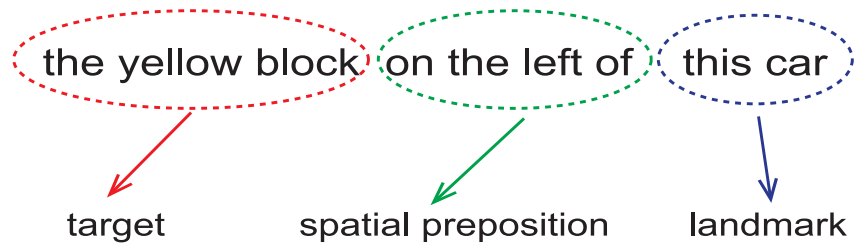

Fig. 2. Components of a spatial relation unit in a referring expression.

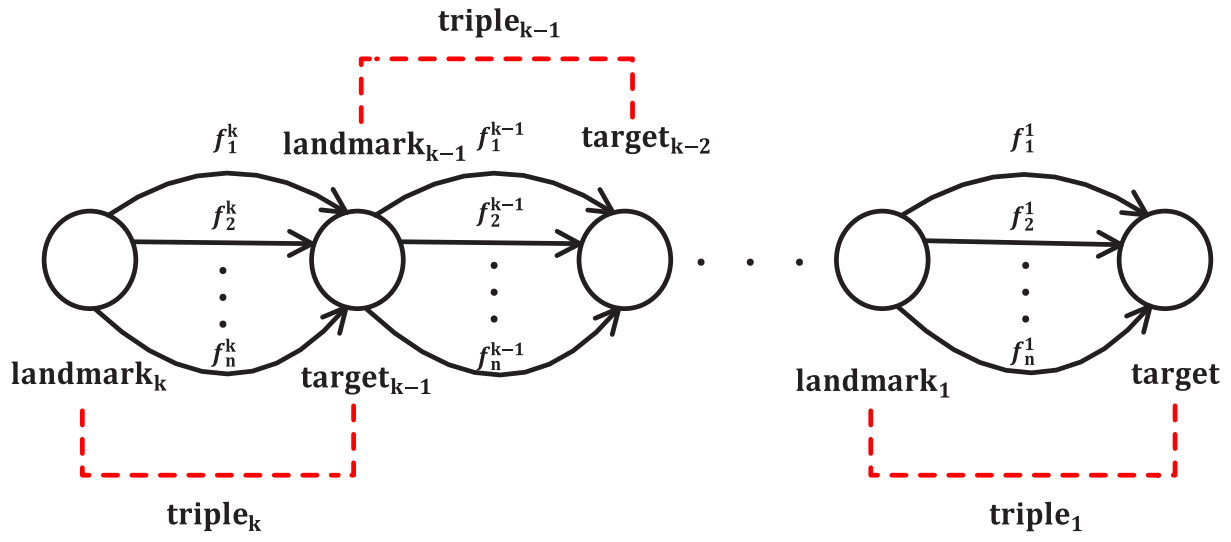

Fig. 3. Diagram of the REG problem.

The robot's task is to generate a referring expression (e.g., "the block on the left of this car") and refer an entity to humans. Given a particular interactive scenario and an intended target entity $o_t$, the robot should output an appropriate expression RE. The input and output can be symbolized as

$$\begin{aligned} \text{Input} &= \{\mathcal{O} = \{o_1, o_2, \ldots, o_d\},\ o_t\} \\ \text{Output} &= \text{RE} \end{aligned} \tag{1}$$

where $\mathcal{O}$ is the set of the robot's world model of entities. It is clear that there are three components in the sentence through dissecting the spatial referring expression (as shown in Fig. 2), which are the landmark $o_l$, the target to be referenced $o_{\text{target}}$, and the spatial preposition $r_{sp}$, respectively. In this article, we represent the sentence as a triple of $(o_l, r_{sp}, o_{\text{target}})$ to characterize an independent spatial relation unit. Actually, an expression may contain more than one spatial relation unit. We assume that the generated expression RE consists of $k$ relation units, where $k$ reflects one aspect of the complexity of RE, for the sake of simplicity, we define the complexity of RE as $k$ in this article. For example, for the expression of "the yellow block on the left of this car" $k = 1$ while for "the red triangle in front of the cuboid on my left" $k = 2$.

The diversity of reference frames means that humans can take any one of the reference frames to resolve the spatial preposition. From this point of view, the spatial prepositions are tightly bound together with reference frames. For robots, spatial prepositions can only be generated if the reference frame is selected in advance. Assume that there are $n$ reference frames and REG can be abstracted as the process of combining graph generation and path searching (as shown in Fig. 3). In the $i$th spatial relation unit, we use $f_j^i$ to represent the corresponding reference frame. We call a combination of reference frames as a reference frame strategy

$$\tau = \left\{f_{j_1}^1, f_{j_2}^2, \ldots, f_{j_k}^k\right\} \tag{2}$$

where $f_{j_i}^i \in \mathcal{F}$, $i = 1, 2, \ldots, k$, and $\mathcal{F}$ is the set of reference frames.

A number of factors have been identified that influence the preference of choosing a particular reference frame [41], such as the relationship between the targets and landmarks, characteristics of the entities in the scene, the communicative purpose of the task, and previous discourse. We believe that the real preference of reference frames can be observed when all these factors are held constant. We assume that the estimated preference of each frame is $p_{f_j}$, $j = 1, 2, \ldots, n$. If the nodes shown in Fig. 3 are ignored, robots will be demanded to take a best reference frame strategy $\tau^*$ based on the observed preference to generate an expression RE to minimize the ambiguity of spatial referring expression for humans.

## B. Framework

As shown in Fig. 3, we can characterize REG as the process of combining graph generation and path searching. Intuitively, the first step is to identify the landmark $o_l$. Specifically, we utilize a modified locative incremental algorithm (M-LIA) to circularly generate landmarks used in the expression to locate the target. In general, the target $o_{\text{target}}$ is identified in advance, hence the next step is to generate spatial prepositions to reflect the spatial relationship between $o_{\text{target}}$ and $o_l$. The biggest obstacle in this stage is that there is a strong correlation between spatial prepositions and reference frames, and the spatial preposition can only be generated if the reference frame is determined. We need to select the most appropriate reference frame for every triple to generate a spatial referring expression that can be understood most easily. As a matter of fact, the selection of reference frames is related to different spatial relation units, and we should consider a whole reference frame strategy rather than selecting a reference frame independently in every spatial relation unit.

To annotate every referring expression with a measure of how easy it is for the listener to resolve, we propose a probabilistic referring expression resolution model based on the comprehension strategy of a literal listener [42]. This model captures the probability that the listener will resolve a given expression RE to the entity in this domain. Finally, we find the best referring expression that optimizes the appropriateness and effectiveness in a joint by enumerating all possible expressions of bounded complexity.

The whole process is based on the assumption that the real preference of reference frames can be observed, while it is unrealistic. In order to solve this problem, we use estimated preference probabilities to approximate the real preference probabilities. Instead of generating a complete referring expression for the target object at once, we provide an iterative mechanism for generating expressions to adapt to the update of the estimated preference probabilities. The framework of PcSREG is summarized as in Algorithm 1 and illustrated by

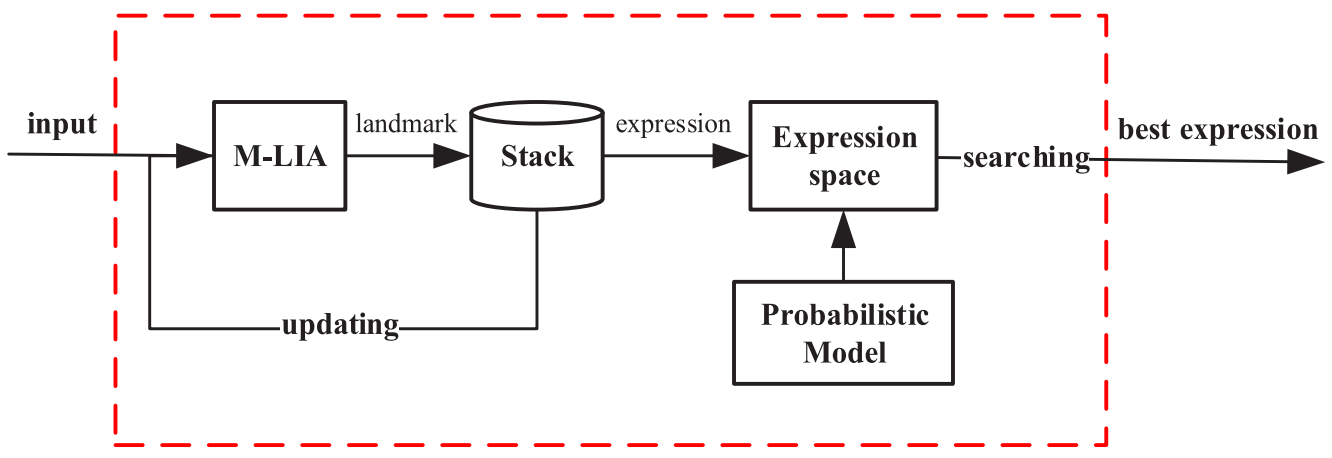

Fig. 4. Block diagram of PcSREG.

**Algorithm 1:** Framework of PcSREG

1. Get initial preference probabilities for reference frames selection;
2. Select all landmarks using the M-LIA algorithm (*Algorithm* 2) and store them in the stack model;
3. Update preference probabilities and go to Step 2 until the preference probabilities remain unchanged;
4. Generate expression space based on the selected landmarks and all possible reference frame strategies;
5. Utilize the probabilistic model to evaluate every referring expression in the expression space and pick out the best one.

the structure diagram as shown in Fig. 4, which will be implemented with specific techniques and algorithms introduced in the following subsections.

## C. Landmark Selection

The landmark selection is implemented using an M-LIA, which can effectively address the combinatorial explosion issue.[1] Although landmark selection is a subtask of REG, the process of landmark selection still needs to cover all the REG steps except the generation of spatial prepositions. The M-LIA is shown as in Algorithm 2.

In the original locative incremental algorithm [31], there is a prevailing assumption that one should only use a spatial reference when the visual features are not sufficient to distinguish the intended target $o_t$. But this assumption is in conflict with the experimental fact that spatial relations are often used even though they are not strictly required [43]. In this article, we aim at helping robots generate a referring expression that is understood as intended by humans as far as possible. As studies show that using spatial references in the expression has a significant negative effect on the comprehension accuracy of human [14], it is reasonable to use spatial reference in expressions as a last resort. Hence, as shown in Algorithm 2, after initializing a default reference frame $f$ randomly (line 1), we preferentially search visual features to describe the target utilizing the basic incremental algorithm, like locative incremental algorithm [31] (line 2). If it fails to generate a distinguishing expression using visual features, we have to use spatial references to locate the target by selecting landmarks (lines 6–19). We modified the original locative incremental algorithm considering the priority of candidate landmarks and the constraints on landmark selection.

[1]The construction of the graph containing all the relationships between all the entities is prone to combinatorial explosion in GBA, hence our proposed method is based on an extended incremental algorithm. However, our method can also be reformulated as a GBA method.

*Remark 1:* Spatial preposition models for different reference frames with 90° rotations respect to each other were proved to be very similar [41], and the reference frames used in our article are always with rotations being a multiple of 90° with respect to the others. Hence, selecting a default reference frame randomly (as shown in line 1, Algorithm 2) will have no effect on the generation of landmarks.

In M-LIA, first, instead of ordering the candidate landmarks through salience (including visual salience and discourse salience), we utilize the entropy of the preference distribution to set the priority of the candidate landmarks (lines 7 and 8). The entropy of preference distribution is used to indicate how disordered the reference frames are and is denoted as

$$H = -\sum_{j=1}^{n} p_{f_j} \times \lg p_{f_j} \tag{3}$$

where $f_j \in \mathcal{F}$, $j = 1, 2, \ldots, n$. The greater the entropy $H$ is, the higher the uncertainty will be. We should avoid selecting the objects with high uncertainty as landmarks. Second, we modified the constraints on the landmark selection. Spatial prepositions are ambiguous inherently and there may be two or more entities sharing the same spatial relationship with the target in the case of adopting a rigid spatial preposition model. Hence, the basic constraints on the landmark selection are that the spatial relationship between the landmark and the target is true and this relation should be different to other relations between this landmark and other distractors (lines 12 and 13). Besides, instead of generating landmarks and spatial prepositions at the same time, we only select suitable landmarks (lines 10–18) under the default reference frame $f$.

## D. Landmark Updating

As a matter of fact, it is hard to observe the real preference of reference frames. The real preference of reference frames is related to the context of expression, but we cannot foresee the expression until we actually generate it. Under this circumstance, we use an initial estimated preference probability $p_{f_j}^0$ $(j = 1, 2, \ldots, n)$ to approximate the real preference probability, so that we can select the landmarks without hindrance using Algorithm 2.

In a real-life scenario, the complexity $k$ is usually greater than one. Hence, we propose a stack model to store the selected landmarks as shown in Fig. 5. The stack model not only helps us to adapt to the update of the estimated preference probability, but also is conducive to generating spatial prepositions. Instead of generating a complete referring expression immediately, we select appropriate landmarks according to the current estimated preference probability $p_{f_j}^c$ and store them in the stack. Then, there is an assumption that we can get an updated preference probability $p_{f_j}^{c+1}$. If $p_{f_j}^{c+1} \neq p_{f_j}^c$, we will select new landmarks according to $p_{f_j}^{c+1}$ and cover the last stack. We perform the above process repeatedly until the estimated preference probability remains unchanged. Finally, we

**Algorithm 2:** M-LIA Algorithm

**Input**: $\mathcal{O} = \{o_1, o_2, \ldots, o_d\}$; $o_t \in \mathcal{O}$; $p_{f_j}$ for all objects
**Output**: description $d_{vf}$, landmark $o_l$

1 Initialize default reference frame $f$ randomly;
2 Generate description $d_{vf}$ for $o_t$ using visual features;
3 $o_l \leftarrow None$;
4 **if** $d_{vf}$ *is distinguishing* **then**
5 Return $d_{vf}$, $o_l$;
6 **else**
7 $\mathcal{O}_{cl} = \{o | o \in \mathcal{O} \; \& \; d_{vf}(o) = False\}$;
8 Set the priority of candidate landmarks according to entropy $H$;
9 $\mathcal{O}_{dt} = \{o | o \in \mathcal{O} \; \& \; o \neq o_t \; \& \; d_{vf}(o) = True\}$;
10 **for** $q = 1$ *to* $|\mathcal{O}_{cl}|$ **do**
11 **for** $g = 1$ *to* $|\mathcal{R}|$ **do**
12 **if** $Relation(\langle o_t, o_q \rangle | frame = f; \; o_q \in \mathcal{O}_{cl}) = r_g \; \&$ $Relation(\langle o_{dt}, o_q \rangle | frame = f; \; o_q \in \mathcal{O}_{cl}; \; o_{dt} \in \mathcal{O}_{dt}) \neq r_g$ **then**
13 $o_l \leftarrow o_q$;
14 Return $d_{vf}$, $o_l$;
15 **end**
16 **end**
17 **end**
18 **end**
19 Return False.

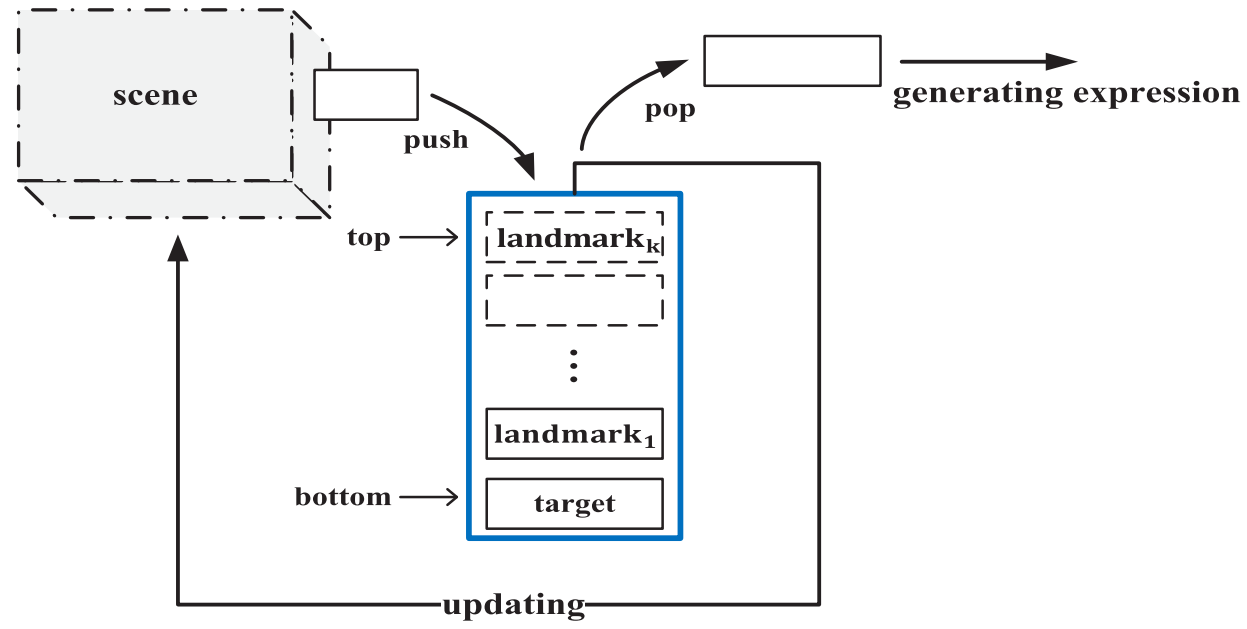

Fig. 5. Stack model for storing selected landmarks.

pop the landmarks out of the stack circularly to generate the complete expression.

### *E. Best Referring Expression Generation*

The last stage of PcSREG is generating the best referring expression using the selected landmarks. There are two important steps: one is to measure each expression by the proposed probabilistic referring expression resolution model and the other is to search for the best solution.

*1) Probabilistic Referring Expression Resolution Model:* Given the referring expression RE in the domain $\mathcal{O} = \{o_1, o_2, \ldots, o_d\}$, we use a probabilistic model $p_{\text{man}}(o|\text{RE})$ to capture the probability that human will resolve RE to the object $o$. By this probabilistic model, we can measure the quality of the generated expression. In this article, we parse the expression by syntax tree derived from context-free grammar. Specifically, each code in the parse tree has a denotation, which is computed recursively in terms of the node's children via a set of simple rules. The denotation is the probability distribution over all entities in the domain.

We define the context-free grammar as $G = (V, \Sigma, W, S)$, where $V$ is a finite set of nonterminal nodes, $\Sigma$ is a finite set of terminals that make up the actual content of the expression, $W$ is the set of rewrite rules, $S \in V$ is a start symbol. The set of nonterminal nodes $V = \{\text{NP}, \text{PP}, \text{IN}\}$. Different from general grammar, we consider the expression that conform to noun phrase (NP), prepositional phrase (PP), or preposition or conjunction (IN). We ignore the verb phrase (VP) to focus on referring the target rather than communicating operational instructions to the listener. Besides, to simplify the calculation, we get rid of some basic grammatical categories, including noun ($N$), article (AT), and adjective (JJ). These basic grammatical categories can be included in a NP without being further parsed. For example, "the red block" can be parsed as a basic NP. The set of rewrite rules $W$ is shown as follows:

$$
\begin{aligned}
S &\rightarrow \text{NP} \\
\text{NP} &\rightarrow \text{NP PP} \\
\text{PP} &\rightarrow \text{IN NP}.
\end{aligned}
\tag{4}
$$

Given a subtree $\Lambda$ rooted at $v \in \{\text{NP}, \text{PP}\}$, we define the denotation $P(\Lambda)$ to be a probabilistic distribution over the entities in the domain. The probabilistic resolution model $p_{\text{man}}(o|\text{RE})$ is computed recursively as follows.

1) If $\Lambda$ is rooted at NP with a single child $x$, then $P(\Lambda)$ is the uniform distribution over $\Psi(x)$, where $\Psi(x)$ is the set of entities consistent with the phrase $x$ [42]. $\mathbb{I}[o \in \Psi(x)] = 1$ if $o \in \Psi(x)$, and 0 otherwise.
2) If $\Lambda$ is rooted at NP with multiple children, then we recursively compute the distribution over entity $o$ for each child tree, multiply the probabilities, and renormalize [42]. Subjecting to the rewrite rules proposed above, there are no more than two child trees.
3) If $\Lambda$ is rooted at PP with a spatial preposition $r$, we recursively compute the distribution over entity $o'$ for the child NP tree. We then appeal to the base case to produce a distribution over entity $o$ related to $o'$ via the spatial preposition $r$. In this article, we emphasize the fact that spatial preposition depends on the particular reference frame. For example, given "the object in front of the square" in the scene shown as in Fig. 6 (a), object $A$ will be identified as the target according to the perspective of the listener and object $D$ will be identified according to the perspective of the speaker. Hence, we should compute the probability $p_{\text{man}}(o|\text{RE})$ by multiplying the preference of each reference frame.

These rules are defined formally as follows:

$$
p_{\text{man}}(o|\text{RE}) \propto \begin{cases} \mathbb{I}[o \in \Psi(x)], & \Lambda = (\text{NP } x) \\ \prod_{q=1}^{g} p_{\text{man}}\big(o|\Lambda_q\big), & \Lambda = \big(\text{NP } \Lambda_1 \cdots \Lambda_g\big) \\ \sum_{o', f_j} p_{\text{man}}\big(o|(r, o')\big) p_{f_j} p_{\text{man}}\big(o'|\Lambda'\big), & \Lambda = \big(\text{PP (IN } r) \; \Lambda'\big). \end{cases}
\tag{5}
$$

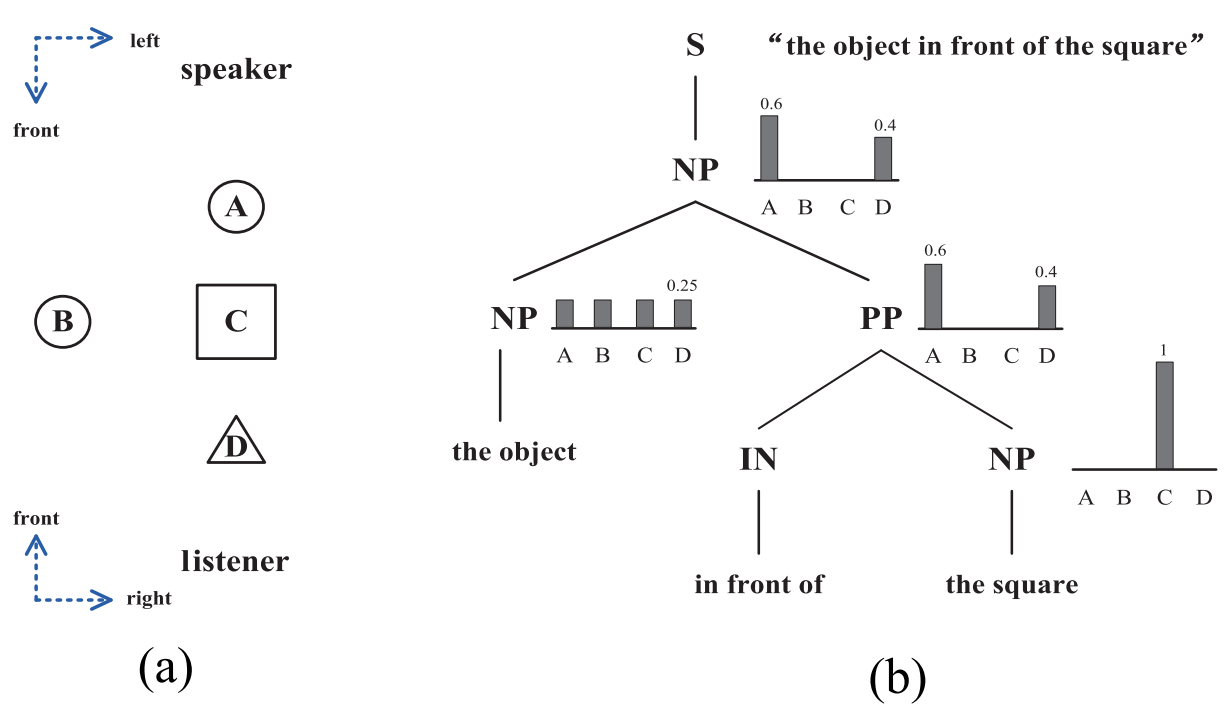

Fig. 6. Example of HRI using the probabilistic referring expression resolution model. (a) Speaker and the listener stand opposite, and there are four objects in the scene. (b) Parse tree modeling the process of listener resolves an expression.

*Example 1:* Supposing that there are only two reference frames (speaker's perspective and listener's perspective) as shown in Fig. 6(a), and their real preference probabilities are $p_{f_1} = 0.4$ and $p_{f_2} = 0.6$ for the listener. Fig. 6(b) shows an example of this bottom-up denotation computation for the expression "the object in front of the square." The denotation starts with the lowest NP node "the square," there is only an object $C$ consistent with this phrase, so the probability that $C$ is identified is 1. Moving up the tree, we compute the denotation of the child tree PP. The listener will take his perspective with a probability of 0.6 to resolve the spatial preposition "in front of," under this circumstance, we can get the distribution $0.6 \times \{1, 0, 0, 0\}$. On the contrary, we can get the distribution $0.4 \times \{0, 0, 0, 1\}$ by taking the perspective of the speaker. The result is the sum of these two parts. The denotation of the node "the object" is a flat distribution over all the objects in this domain. Finally, the denotation of the root is computed by taking a product of the object distributions and turns out to be the same split distribution as its PP child.

*2) Best Referring Expression Generation:* The best referring expression should satisfy two conditions, i.e., appropriateness and effectiveness. Given an expression RE for target $o_t$, we use a binary function $\omega_1(\mathrm{RE})$ to represent the appropriateness

$$\omega_1(\mathrm{RE}) = \begin{cases} 1, & p_{\mathrm{man}}(o_t|\mathrm{RE}) = \max\{p_{\mathrm{man}}(o|\mathrm{RE})|o \in \mathcal{O}\} \\ 0, & \text{otherwise.} \end{cases} \tag{6}$$

The appropriateness determines the fit of an expression for the target in this domain. If $p_{\mathrm{man}}(o_t|\mathrm{RE})$ is not the biggest of all, it shows that there is another object better suited to this expression than the target $o_t$. It is inappropriate to view this RE as the best expression. As for effectiveness, we characterize it with the probability that the listener will resolve RE to the target $o_t$

$$\omega_2(\mathrm{RE}) = p_{\mathrm{man}}(o_t|\mathrm{RE}). \tag{7}$$

The higher the effectiveness $\omega_2(\mathrm{RE})$ is, the higher the accuracy of the interpretation will be. To generate the best expression, we should optimize both of the appropriateness and effectiveness in a joint although they are unified in general. For the robot to generate the best referring expression, we define the strategy with the probability that the robot generates spatial description RE for target $o_t$

$$p_{\mathrm{rob}}(\mathrm{RE}|o_t) = \begin{cases} 1, & \mathrm{RE} = \mathrm{RE}^* \\ 0, & \mathrm{RE} \neq \mathrm{RE}^* \end{cases} \tag{8}$$

where $\mathrm{RE}^* = \arg\max_{\mathrm{RE}'} [\omega_1(\mathrm{RE}') + \omega_2(\mathrm{RE}')]$.

As shown in Fig. 3, the solution space is reduced to a combination of reference frames, i.e., reference frame strategy $\tau$. We perform this maximization by exhaustive search to guarantee its convergence and optimality. To be specific, we pop the selected landmarks out of the landmarks stack circularly and generate all possible referring expressions according to all possible reference frame strategies. We call the set of the possible referring expressions as an expression space. Finally, we traverse the expression space to find the best expression.

**Algorithm 3:** PcSREG Algorithm

**Input**: The model of the world $\mathcal{O} = \{o_1, o_2, \ldots, o_d\}$; the target entity $o_t \in \mathcal{O}$; the initial preference probability $p_{f_j}^0$ for reference frame $f_j \in \mathcal{F}$

**Output**: The best referring expression RE*

1 $p_{f_j} \leftarrow p_{f_j}^0$;
2 **while** $p_{f_j}^{c+1} \neq p_{f_j}^c$ **do**
3 $\quad \mathcal{O}' \leftarrow \mathcal{O}$, $o_t' \leftarrow o_t$, $D_{vf} \leftarrow \{\}$, *Stack* $\leftarrow \{\}$, $d_{vf} \leftarrow \mathit{None}$;
4 $\quad$ **while** $d_{vf}$ *is not distinguishing* **do**
5 $\quad\quad [d_{vf}, \mathrm{o}_l] = \text{M-LIA}(\mathcal{O}', o_t', p_{f_j})$ using *Algorithm* 2;
6 $\quad\quad D_{vf} = D_{vf}.\mathrm{append}(d_{vf})$;
7 $\quad\quad \mathit{Stack} = \mathit{Stack}.\mathrm{push}(\mathrm{o}_l)$;
8 $\quad\quad \mathcal{O}' \leftarrow \mathcal{O}' - \{o|o \in \mathcal{O}' \ \& \ d_{vf}(o) = \mathit{True}\}$;
9 $\quad\quad o_t' \leftarrow \mathrm{o}_l$;
10 $\quad$ **end**
11 $\quad$ Re-assess the preference probability $p_{f_j}^{c+1}$ of reference frames;
12 $\quad p_{f_j} \leftarrow p_{f_j}^{c+1}$;
13 **end**
14 Pop the landmarks out of *Stack* and generate the expression space;
15 Search for the best reference frame strategy $\tau^*$ using the probability referring expression resolution model; Generate the optimal set of relation prepositions $D_{sp}$ according to $\tau^*$;
16 $\mathrm{RE}^* \leftarrow D_{vf} \cup D_{sp}$;
17 Return RE*.

## *F. Integrated PcSREG Algorithm*

The integrated PcSREG algorithm is shown as in Algorithm 3. First, with initial preference probabilities, we reuse M-LIA (Algorithm 2) to generate landmarks that help to locate the target entity until the visual features can be used to distinguish the current landmark and push them in the stack model (lines 4–10). Second, after reassessing the preference probabilities, we will regenerate the landmarks and update the Stack (lines 2–13), and this process will not end until the preference probabilities remain unchanged. Third, we pop all landmarks out of the stack and generate the expression space (line 14). Furthermore, with the probabilistic model, we find the best reference frame strategy $\tau^*$ (line 15). Finally, we generate the optimal set of relation prepositions according to $\tau^*$ (line 16) and integrate the description of all landmarks including the target entity and optimal relation prepositions to generate the final expression RE* (line 17).

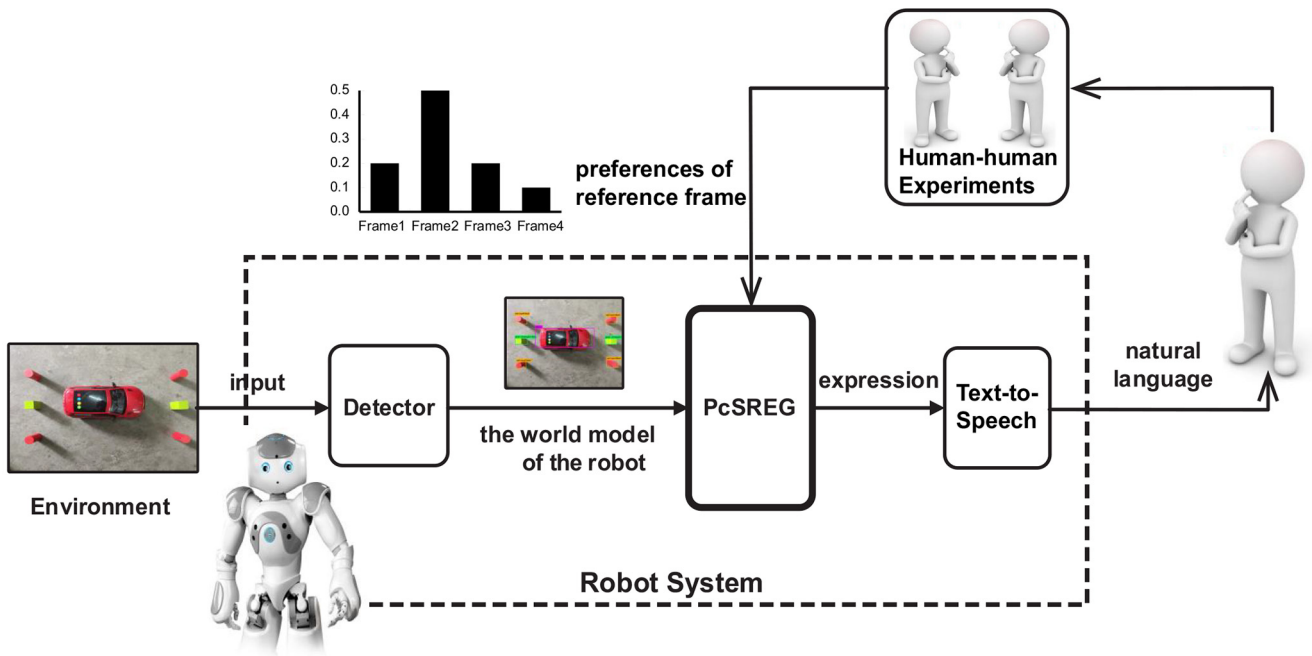

Fig. 7. Architecture of the implemented robot system with PcSREG.

*Remark 2:* PcSREG guarantees that the generated expression is optimal. Although multiple objects can be functioned as the landmarks, we use the entropy of preference distribution to set the priority in Algorithm 2. Furthermore, the exhaustive search method can be proved to find the best reference frame strategy.

*Remark 3:* As for the complexity of PcSREG, in the landmark selection procedure, it would not consume any computation for extra variables so that Algorithm 2 is bounded by the same computational complexity as the locative incremental algorithm [31]. Suppose that the updating process described in Section III-D will converge after $l$ steps, we need to multiply the computational complexity with $l$. In the reference frame strategy selection procedure, let $n = |\mathcal{F}|$, the computational complexity of searching the best strategy is $O(n^k)$ for generating a referring expression with complexity $k$. As a result, a large $k$ will greatly increase the computational complexity of our method.

*Remark 4:* Real-world interaction scenarios are quite complex, and the performance of the proposed algorithm is strongly related to the appropriateness of the spatial preposition model. The proposed PcSREG algorithm is based on the projective preposition model, which will be described in detail in Section IV-B.

## IV. System Implementation

To evaluate the proposed approach, we implemented PcSREG on a robot system, whose architecture is shown as in Fig. 7. The PcSREG approach depends on the preference of reference frames, hence we conduct human–human interactive experiments to estimate the preference and transport it to the PcSREG module. The process of human–human experiments is described in Section V-A. As shown in Fig. 7, first, the robot perceives the environment and converts environmental information into its internal representation. Then, the robot generates the optimal referring expression RE* using Algorithm 3 and finally, the robot turns RE* into natural language for the human partner to refer the target entity $o_t$ successfully.

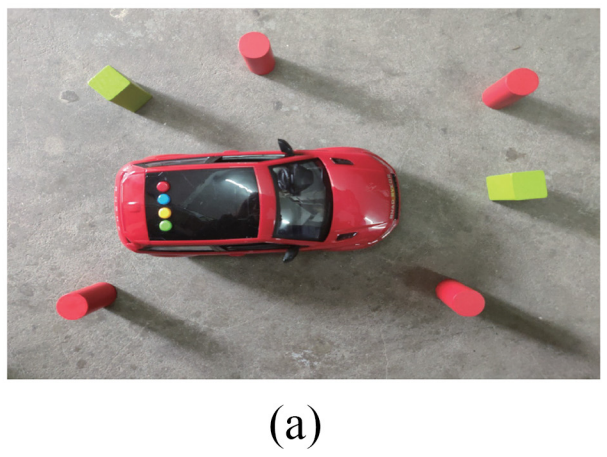

(a)

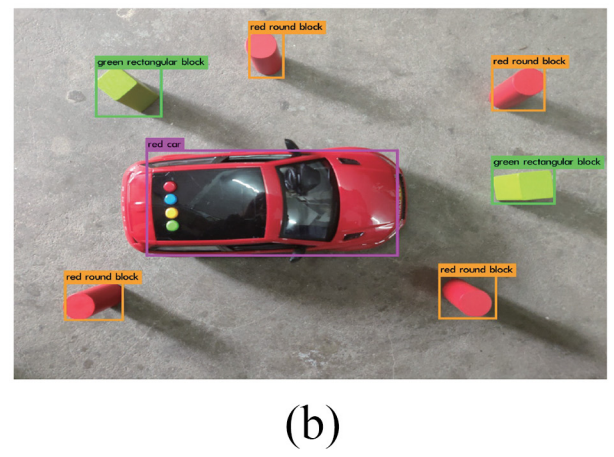

(b)

Fig. 8. Designed interactive scene. (a) Sample of the collected data set. (b) Detection results.

### A. Object Detection

To detect the objects in a scene, we use YOLOv3, one of the mainstream, real-time object detection system based on residual networks [44]. YOLOv3 is extremely fast and accurate, and it is a one-stage object detection method that is different from R-CNN. Specifically, YOLOv3 applies a single neural network to the full image. This network divides the image into regions and predicts bounding boxes and probabilities for each region. These bounding boxes are weighted by the predicted probabilities.

In this article, we create a data set of the tabletop task for training our object detector. This data set only contains 100 samples and it consists of toy cars and various blocks as shown in Fig. 8(a). We use the tool Yolo_mark[2] that is a GUI for marking bounded boxes of objects in images to label samples. Because of the small data set, our YOLOv3 model uses a small model yolov3-tiny[3] as the entrained model. As a result, we can get the visual features and locations of all objects in the image as shown in Fig. 8(b), for example.

### B. Spatial Preposition Model

From a geometric semantic point of view, spatial prepositions can be divided into two classes: topological prepositions (e.g., "near") and projective prepositions (e.g., "front") [45]. Projective prepositions depend on the selected reference frame while topological prepositions do not. In addition, the effectiveness of referring expression is strongly related to the performance of the preposition model. In this article, we model the projective prepositions solely so that we can avoid distractions outside the reference frames.

It is reasonable that spatial prepositions are represented as fuzzy sets to capture the ambiguity inherent to the linguistic terms for the prepositions [46]. Hence, bounding boxes are appropriate representations for objects of all possible shapes. In this case, the centroids of the objects are used in evaluating the fuzzy membership functions of the projective relations.

Considering that projective prepositions are conditioned on reference frames, so spatial preposition models should not be independent of reference frames. It is generally agreed that there are three kinds of reference frames: 1) intrinsic reference frame; 2) relative reference frame; and 3) extrinsic reference frame, whose implications will be explained in Section V-A. To simplify the models, we specify that the canonical axis for "front" is consistent with the origin's view direction in both intrinsic and relative reference frames. It is shown that spatial preposition models for different reference frames are very similar [41], hence we just need to model the spatial prepositions for a particular reference frame and model the spatial prepositions through changing the direction of the reference axis to adapt to other reference frames.

[2]https://github.com/AlexeyAB/Yolo_mark
[3]https://pjreddie.com/media/files/yolov3-tiny.weights

### C. Linguistic Realisation

Actually, there are two questions that need to be answered in REG, i.e., 1) which set of properties distinguishes the target (the content selection) and 2) how the selected properties are to be turned into natural language (linguistic realization) [9]. In our work, we mainly focused on the former and use a simple sentence template to turn the selected contents into text, then we utilize the speech synthesis API provided by Baidu AI Open Platform[4] to convert this text into natural language.

## V. Experiments

To generate an effective referring expression via PcSREG, we contribute a study of human–human interaction to estimate the preference of reference frames. In the study, we collect a corpus of spatial references generated by participants. Then, we code reference frames manually for the collected corpus and estimate the probabilities from a statistical point of view. The probabilities include the initial probabilities and the updated probabilities that takes the content of expression into account. After that, we evaluate the effectiveness of the generated reference expressions in a user study.

### A. Preference Estimation

We conduct a study where participants are asked to give natural language instructions to a partner sitting across the table to pick up an indicated object from the table, and we try to get the preference of reference frames that approximates the real preference as closely as possible so that we can generate effective referring expressions for real scenarios.

*1) Corpus Collection:* Speakers tend to select reference frames differently with actual conversational partners than with the usually studied imaginary addressees [22]. To get more realistic data, we conducted this study in an actual conversational situation rather than online study through crowdsourcing marketplaces [14]. In this case, we created a set of stimulus scenes. Each scene represents a configuration with multiple objects in different colors and shapes on a table, besides different scenes contain different numbers of objects and the arrangements of objects are inconsistent (see *Supplementary Material* for more details). This stimulus design is chosen to elicit referring expressions that rely more on the projective prepositions depending on the reference frame. During the study, in order to capture a clear referring expression, participants are asked to instruct a partner (i.e., an experiment organizer) sitting across the table to pick up the referenced entity. The referenced entity is pointed out by an organizer beforehand and the expressions given by participants are recorded immediately. Throughout the whole process, participants are asked to generate expressions for eight trails, there is only one intended target in each trial. At the end of this study, we collected demographics of every participate, such as age and gender.

*2) Reference Frame Metrics:* A reference frame may include an origin, a coordinate system, a point of view, terms of reference, and a reference object [23], and we may take it as an abstract coordinate system simply. By this definition, three kinds of reference frames depending on their origins are presented, i.e., relative, intrinsic, and extrinsic. In relative uses, the origin of the coordinate system is one of the participants, the speaker (egocentric perspective) or the listener (addressee-centered perspective), which respectively correspond to the robot and human in this article. In intrinsic uses, the origin of the coordinate system is a specific object with orientations. In this case, we do not require that the specific object must be functioned as a landmark. In extrinsic uses, the origin of the coordinate system is external to the scene. The most common extrinsic coordinate system is the cardinal directions (north, south, east, and west).

We code the reference frames manually to analyze which category of reference frames is used in the collected reference expressions. As a matter of fact, the expressions generated by participants are more complex than those consisting of one spatial relation unit. Each spatial relation unit depends on a particular reference frame, and we regard each spatial preposition in the expressions as a basic unit to analyze the reference frames. For instance, the expression "the red triangle in front of the cuboid on my left," "the red triangle in front of the cuboid," and "the cuboid on my left" are two basic analytical units. Intuitively, there is a strong correlation between the category of landmarks and the reference frames. For instance, we tend to use the egocentric perspective in the utterance "the block on my left." Inspired by this, instead of coding the reference frame for each spatial relation unit, we code the used reference frame for different types of landmarks (speaker, listener, objects with orientations, and objects without orientations) according to the actual conversational situations. Particularly, when the keywords, such as "north," "south," "east," and "west" appear in the expression, we consider the perspective to be extrinsic (see Table II for details).

*3) Statistical Results:* In this study, we recruited 130 participants (75 Males, 55 Females) and oversampled 933 expressions so that we could account for errors in the data collection process and invalid responses. The ways in which participants reference objects vary, we remove 524 sentences that either do not involve any reference frame (96%) or are otherwise nonsensical (4%). For instance, the expression of "the cuboid near the car" does not involve any reference frame. As a result, there are 463 valid basic units in total, and we analyze these basic units for reference frames.

First, we analyzed the preference of reference frames in the most intuitive way. We simply counted usage ratios of all reference frames for different types of landmarks according to the collected corpus that has been coded. The statistical results are shown as in Table III. We conducted a Chi-square ($\chi^2$) test of independence between the landmark type and the used reference frame, the hypothesis that there are significant relationship between the two variables is verified with $\chi^2(9) = 527.706$, $n = 463$, $p < 0.001$. We treat this result as the initial preference probabilistic distribution $P^0$ for selecting appropriate reference frames. For instance, it means that when the listener is selected to function as a landmark by the robot, the listener will take the robot's perspective (i.e., egocentric perspective) with a probability of 4.08% to resolve the corresponding spatial preposition and take his own perspective (i.e., addressee-centered perspective) with a probability of 95.92%.

[4]https://ai.baidu.com/tech/speech/tts

TABLE II
POSSIBLE REFERENCE FRAMES

| Landmarks | Utterances | egocentric | addressee-centered | intrinsic | extrinsic |
|---|---|---|---|---|---|
| speaker | "the block on my left" | + | − | − | − |
| listener | "the block on your right" | − | + | − | − |
| objects with orientations | "the block in front of car" | − | − | + | − |
| objects without orientations | "the block in front of the cuboid" | + | − | − | − |
| * | "the block on the north of *" | − | − | − | + |

TABLE III
INITIAL PROBABILITY DISTRIBUTION OF REFERENCE FRAMES

| Landmarks | egocentric | addressee-centered | intrinsic | extrinsic | nums |
|---|---|---|---|---|---|
| **speaker** | 100% | 0% | 0% | 0% | 21 |
| **listener** | 4.08% | 95.92% | 0% | 0% | 98 |
| **objects with orientations** | 4.5% | 4.5% | 90.5% | 0.5% | 200 |
| **objects without orientations** | 66.67% | 20.14% | 11.81% | 1.38% | 144 |

The initial probabilities are rough estimations of real preference, and we try to look for other clues to reveal the preference of reference frames more accurately. A closer look at Table III shows that there is not a reference frame that will be taken with a probability of approaching 1 when the landmark fell into the category "objects without orientations," which means that there is more uncertainty in this case. The content is an important resource in natural language processing and we hypothesize that the real preference is content dependent when the landmark is "objects without orientations." To test this hypothesis, we pick out these expressions whose complexity $k > 1$ and contain "objects without orientations" landmarks, and measure the consistency of reference frames used for different spatial relations in the same expression. The results show that there are 71.88% (23/32) expressions whose reference frames are consistent. To simplify representation, we define content window $[-a, b]$ to represent the context boundary, where $-a$ means that we take the closest $a$ spatial relation units to the left of the current spatial relation unit into account and $b$ specifies the farthest right boundary. Because the complexity $k = 2$ for almost all the selected expressions, we use the context window $[0, 1]$ to couple the content. We have the updated preference distribution of "objects without orientations" landmarks

$$P^{c+1}_{\text{center}} \leftarrow P^{c}_{\text{right}} \tag{9}$$

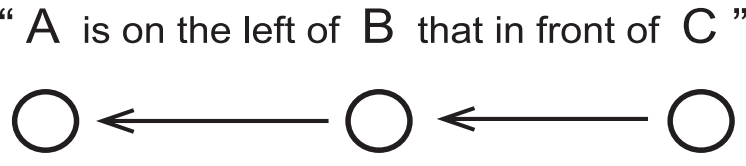

Fig. 9. Example for convergence analysis of landmark updating.

where $P^{c}_{\text{right}}$ is the current preference distribution of the closest landmark to the right.

*Remark 5:* Consider that $k = 2$ and a generated referring expression is shown as in Fig. 9. Assume that the current preference distributions of object $B$ and object $C$ are $P(B)^c$ and $P(C)^c$. Let object $B$ be "objects without orientations," then we will update the preference distribution of object $B$ follows the (9) such that $P(B)^{c+1} \leftarrow P(C)^c$. Let $H(B)$ be the entropy of preference distribution of object $B$, for any $C$, $H(B)^{c+1} \leq H(B)^c$ according to the statistical results in Table III. Hence, the priority of the object $B$ will increase and it is determined that object $B$ will be selected as a landmark again in next updating episode. As a result, the updating process described in Section III-D will converge after at most one step updating. In general, this convergence can be generalized to $k > 2$.

### B. User Study

In user study, to better evaluate the effectiveness of PcSREG, the comparison with various baselines are presented instead of the existing subjective and objective analysis [47].

*1) Experimental Settings:* We aim at minimizing the ambiguity of generated referring expressions for human partners through finding the best reference frame strategy. The proposed PcSREG approach is compared with several baseline methods, i.e., perspective-constant method (including perspective-robot method and perspective-human method), perspective-random method, and PcSREG-max method, whose settings are listed as follows.

1) *Perspective-robot Method:* REG using the egocentric perspective of robot based on the locative incremental algorithm [31]. In HRI situation, the landmark may be human (listener) or robot (speaker), we randomly assign visual salience for them and ensure that they have higher priorities to be selected as a landmark than other objects.

In this setting, we always take the perspective of the robot to execute the task of REG.

2) *Perspective-human:* Same as perspective robot method, the only difference is that we always take the perspective of the human (addressee-centered perspective).

3) *Perspective-random Method:* Generating referring expression according to a randomly selected reference frame strategy.

4) *PcSREG-max Method:* As a more challenging reference point, we report results for a greedy method based on PcSREG. The method discards the probabilistic referring expression resolution model and determines the best reference frame strategy in a greedy way to reveal the superiority of taking the long view for PcSREG, i.e.,

$$\tau^* = \left\{ \arg\max_{f^1_{j_1}} p_{f^1_{j_1}}, \arg\max_{f^2_{j_2}} p_{f^2_{j_2}}, \ldots, \arg\max_{f^k_{j_k}} p_{f^k_{j_k}} \right\}. \quad (10)$$

We conduct all the experiments in real conversational situations rather than online study for real applications. To verify the success of PcSREG to solve the ambiguity of spatial referring expressions, we created a set of experimental scenarios including 12 scenarios numbered as 1–12. Please refer to *Supplementary Material* for more details.[5] In the experiment, participants are asked to interact with the robot face to face. The task of the participants is to point out the referenced entity according to the referring expression generated by the robot (e.g., "the red round block on the left of this car"). Every participant has to be studied in randomly selected four scenarios from all the 12 scenarios, and then in every scene they are asked to resolve five expressions corresponding to the five different methods, respectively. The identified entities are recorded immediately and the demographics of participants (age, gender, and whether the color blindness) are collected at the end of this study.

*2) Experimental Results:* 72 participants are recruited (48 Males, 24 Females) in this study and none of them are color blind. As shown in Fig. 10(a), we measured the accuracy of the target entity identification over all the comparisons. In addition, to test the effectiveness of the proposed method under different complexity, we divided the experimental scenarios into two categories regarding different complexity, i.e., $k = 1$ and $k > 1$. The results are shown as in Fig. 10(b).

*3) Discussion:* As shown in Fig. 10(a), the proposed PcSREG method can achieve the best performance with almost 80% accuracy in the user study. The improvement in human identification accuracy in turn illustrates that PcSREG can effectively reduce the ambiguity of referring expressions. Compared with the perspective-human, perspective-robot, and perspective-random methods, PcSREG takes more attention to the diversity of reference frames and can adaptively adopt appropriate reference frames to generate expression, so the generated expressions by PcSREG are not rigid and can be understood easier. As for PcSREG-max, although it also performs better than the other baselines, it is so short-sighted that it can not find the optimal reference frame strategy. We conducted a McNemar's test and the results show that there are statistically significant difference in accuracy between PcSREG and PcSREG-max with $\chi^2 = 31$, $n = 288$, and $p = 0.002 < 0.01$.

[5]Although there are abundant image datasets (e.g., Google Refexp [39]) can be used in REG, they limit the REG model in plane space which ignores the diversity of reference frames in the real world, hence these image datasets are not suitable for testing our method.

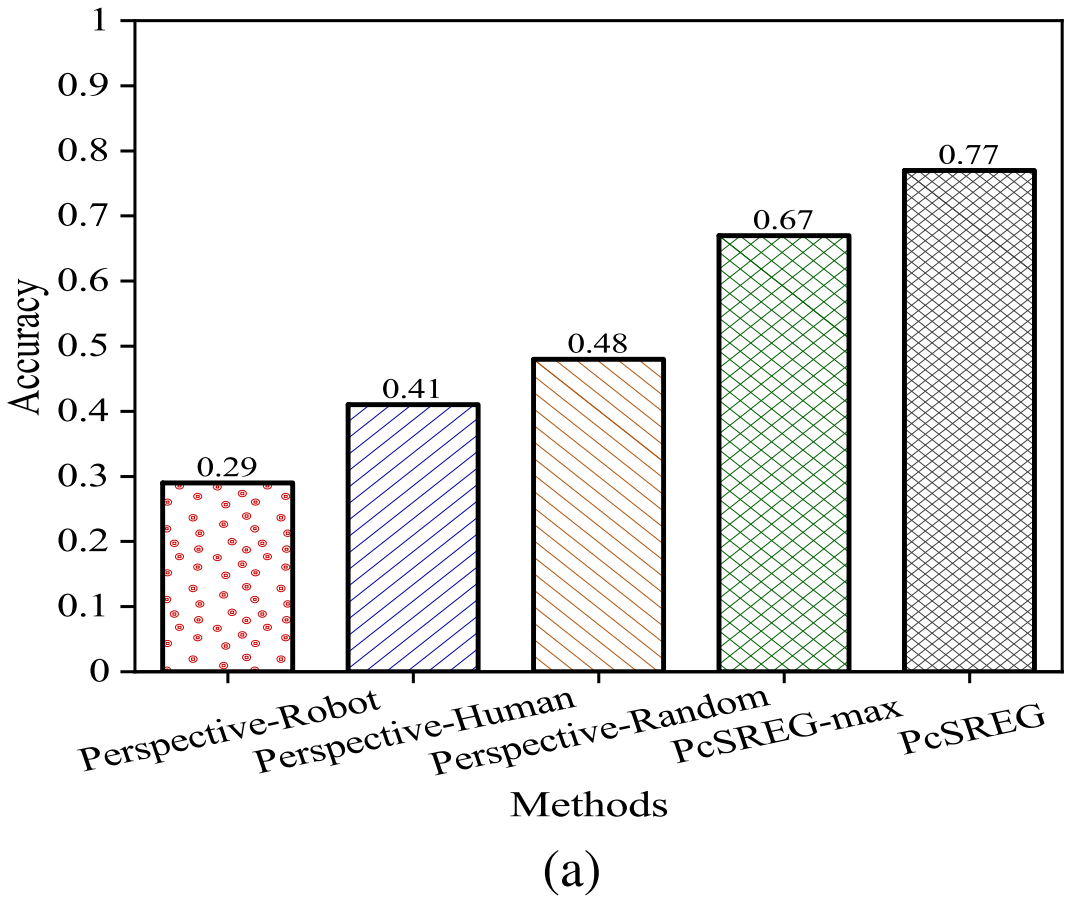

(a)

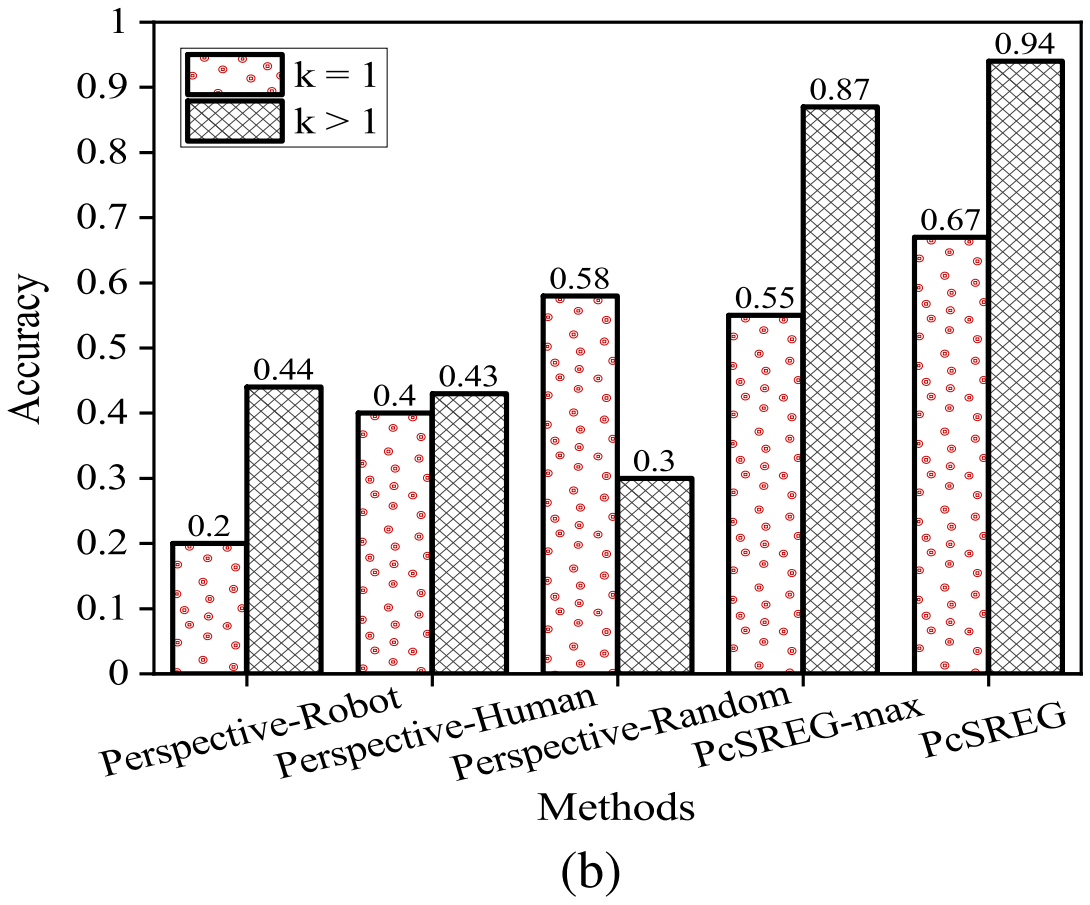

(b)

Fig. 10. Experimental results. (a) Accuracy of target entity identification. (b) Accuracy of target entity identification with different complexity.

Fig. 10(b) shows the accuracy with different complexity $k$. The accuracy with $k > 1$ is higher in most cases. The possible reason is that human does not tend to take the strategy that combines different reference frames to resolve the expressions. According to the perspective-random method, the robot will take different reference frames for different spatial relation units easily, so the generated expressions will be misunderstood easily in the case $k > 1$; while PcSREG can adapt to this characteristic of humans and will prefer to take consistent reference frames for different spatial relation units.

## VI. Conclusion

Our long-term goal is to develop a robot that can interact with human effectively in the real world. In this article, we presented a novel PcSREG framework to minimize the ambiguity of spatial referring expression for human partners by taking the diversity of reference frames into account. We integrated the process of the reference frame selection into the original REG framework. In this way, we can find the optimal reference frame strategy consistent with human preference to facilitate the resolution of humans. In addition, we demonstrated an implementation of PcSREG via a robot system. The experimental results with the preference of reference frames in expression generation and the user study show that the proposed method can effectively reduce the ambiguity of spatial referring expressions for HRI in real scenarios. Our future work will focus on the dynamic interaction and constructing an adaptive interactive robotic system using machine learning methods (e.g., reinforcement learning [48]).

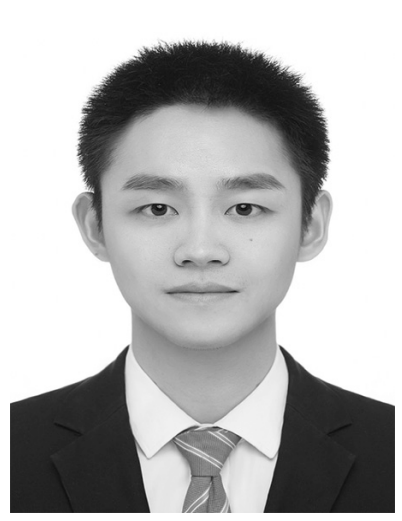

**Mingjiang Liu** received the B.E. degree in automation from the Department of Control and System Engineering, Nanjing University, Nanjing, China, in 2019, where he is currently pursuing the M.S. degree with the Department of Control and Systems Engineering.

His current research interests include human-centered robotics, human–robot interaction, and reinforcement learning.

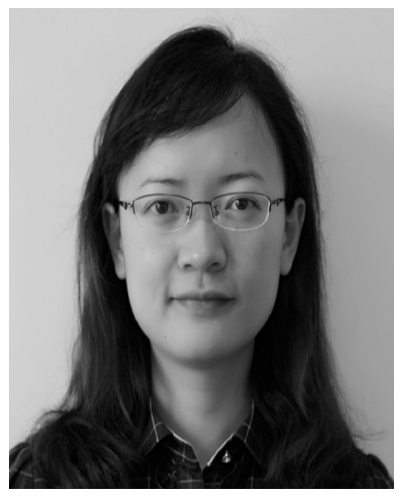

**Chengli Xiao** received the B.E. degree from East China Normal University, Shanghai, China, in 2003, and the Ph.D. degree in psychology from the Chinese Academy of Sciences, Beijing, China, in 2008.

She is currently an Associate Professor and the Associate Head of the Department of Psychology, School of Social and Behavioral Sciences, Nanjing University, Nanjing, China. She was with the Department of Psychological and Brain Sciences, University of California at Santa Barbara, Santa Barbara, CA, USA, from September 2013 to September 2014. Her current research interests include human factors in human–robot interaction and people’s social acceptance of artificial intelligence.

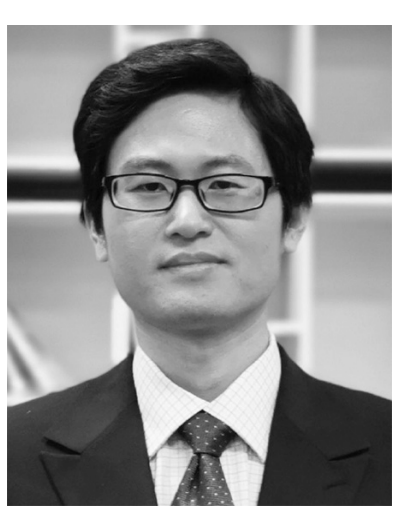

**Chunlin Chen** (Senior Member, IEEE) received the B.E. degree in automatic control and the Ph.D. degree in control science and engineering from the University of Science and Technology of China, Hefei, China, in 2001 and 2006, respectively.

He is currently a Professor and the Head of the Department of Control and Systems Engineering, School of Management and Engineering, Nanjing University, Nanjing, China. He was with the Department of Chemistry, Princeton University, Princeton, NJ, USA, from September 2012 to September 2013. He had visiting positions with the University of New South Wales, Sydney, NSW, Australia, and City University of Hong Kong, Hong Kong. His current research interests include machine learning, intelligent control, and quantum control.

Prof. Chen serves as the Chair of Technical Committee on Quantum Cybernetics, IEEE Systems, Man and Cybernetics Society.